\documentclass[journal]{IEEEtran}

\usepackage{graphicx}%
\usepackage{multirow}%
\usepackage{amsmath,amssymb,amsfonts}%
\usepackage{hyperref}
\usepackage{multirow,tabularx}
\usepackage{amsthm}%
\usepackage{mathrsfs}%
\usepackage{xcolor}%
\usepackage{textcomp}%
\usepackage{manyfoot}%
\usepackage{booktabs}%
\usepackage{algorithm}%
\usepackage{algorithmicx}%
\usepackage{algpseudocode}%
\usepackage{listings}%
\usepackage{adjustbox}
\usepackage[justification=justified]{caption}

\definecolor{level1}{RGB}{230,145,56}
\definecolor{level2}{RGB}{224,102,102}
\definecolor{level3}{RGB}{230,96,165}
\definecolor{level4}{RGB}{61,133,198}
\definecolor{level5}{RGB}{106,168,79}
\definecolor{level6}{RGB}{241,203,90}
\definecolor{greenanalysis}{RGB}{78, 135, 78}
\definecolor{blueanalysis}{RGB}{142,186,217}
\definecolor{yellowanalysis}{RGB}{255,209,127}
\definecolor{greenmaha}{RGB}{106, 168, 79}
\definecolor{yellowmaha}{RGB}{241,194,50}
\definecolor{redmaha}{RGB}{204,0,0}
\definecolor{lightgray}{rgb}{0.83, 0.83, 0.83}
\definecolor{darkgray}{RGB}{94, 94, 94}
\definecolor{blue}{RGB}{0,0,255}

\begin{document}

\title{AI4Food-NutritionFW: A Novel Framework for the Automatic Synthesis and Analysis of Eating Behaviours}

\author{
Sergio Romero-Tapiador\textsuperscript{1,*}, Ruben Tolosana\textsuperscript{1}, Aythami Morales\textsuperscript{1},\\ Isabel Espinosa-Salinas\textsuperscript{2}, Gala Freixer\textsuperscript{2}, Julian Fierrez\textsuperscript{1}, Ruben Vera-Rodriguez\textsuperscript{1}, \\Enrique Carrillo de Santa Pau\textsuperscript{2}, Ana Ramírez de Molina\textsuperscript{2} and Javier Ortega-Garcia\textsuperscript{1}
\\
\textsuperscript{1}Biometrics and Data Pattern Analytics Laboratory (BiDA Lab), Universidad Autonoma de Madrid, Spain\\
\textsuperscript{2}IMDEA Food Institute, CEI UAM+CSIC, Madrid, Spain

*Corresponding author: \href{mailto:sergio.romero@uam.es}{sergio.romero@uam.es} }

\maketitle

\begin{abstract}
Nowadays millions of images are shared on social media and web platforms. In particular, many of them are food images taken from a smartphone over time, providing information related to the individual's diet. On the other hand, eating behaviours are directly related to some of the most prevalent diseases in the world. Exploiting recent advances in image processing and Artificial Intelligence (AI), this scenario represents an excellent opportunity to: \textit{i)} create new methods that analyse the individuals' health from what they eat, and \textit{ii)} develop personalised recommendations to improve nutrition and diet under specific circumstances (e.g., obesity or COVID). Having tunable tools for creating food image datasets that facilitate research in both lines is very much needed.

This paper proposes AI4Food-NutritionFW, a framework for the creation of food image datasets according to configurable eating behaviours. AI4Food-NutritionFW simulates a user-friendly and widespread scenario where images are taken using a smartphone. In addition to the framework, we also provide and describe a unique food image dataset that includes 4,800 different weekly eating behaviours from 15 different profiles and 1,200 subjects. Specifically, we consider profiles that comply with actual lifestyles from healthy eating behaviours (according to established knowledge), variable profiles (e.g., eating out, holidays), to unhealthy ones (e.g., excess of fast food or sweets). Finally, we automatically evaluate a healthy index of the subject's eating behaviours using multidimensional metrics based on guidelines for healthy diets proposed by international organisations, achieving promising results (99.53\% and 99.60\% accuracy and sensitivity, respectively). We also release to the research community a software implementation of our proposed AI4Food-NutritionFW and the mentioned food image dataset created with it. 
\end{abstract}


% INTRODUCTION SECTION

\section{Introduction}
\IEEEPARstart{T}{he} World Health Organization (WHO) estimates that 1.9 billion adults have overweight and more than 600 million of these are obese. As a consequence, many Non-Communicable Diseases (NCD) such as diabetes, cardiovascular, or autoimmune diseases will shortly appear among a significant percentage of the global population \cite{burdenwho}.

\begin{figure*}[t]
    \begin{center}
      \includegraphics[width=1\linewidth]{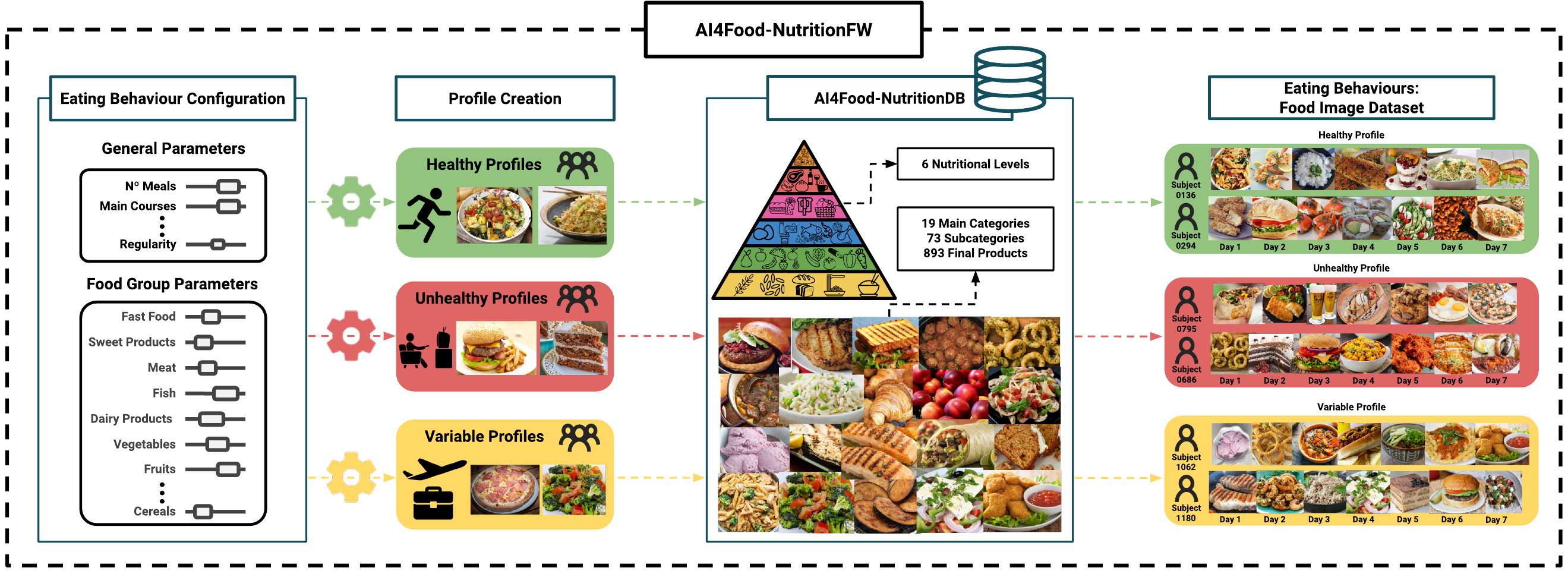}
    \end{center}
      \caption{AI4Food-NutritionFW framework. First, we define the general and food group parameters of our framework to model eating behaviour configurations. Then, by modifying previous parameters, we can define different profiles. Finally, with the help of the AI4Food-NutritionDB \cite{romero}, we can generate food image datasets with different eating behaviours, selecting the number of subjects and profiles as desired, or creating new ones. }
    \label{fig:confenv}
\end{figure*}

Nowadays, meals are composed of a wide range of ingredients and balancing them to reach a healthy diet is a complex and time-consuming task. One solution for generating healthy diets is to follow dietetics guidance from several national and international organisations like nutritional pyramids \cite{ushealth}. However, these are usually general recommendations to the majority of the population and they do not consider individual aspects. Personalised nutrition is expected to be one of the greatest revolutions in this era, thanks in part to the latest advances in Artificial Intelligence (AI). This new trend considers user-specific characteristics like anthropomorphic measures (height, weight, and body composition), genetic factors, gender, personal preferences, physical activity, etc.  \cite{personalisednutrition}, together with some kind of general knowledge or model, to derive user-specific models  \cite{bida12} using statistical approaches or machine learning \cite{bida13}.

Personalised nutrition enables individuals to follow healthier lifestyles and prevent diet-related chronic diseases. As a result, recommendation systems based on healthy diets have recently emerged due to the integration of personalised methods and automatic tools \cite{recommendersurvey}. However, these recommendation systems do not usually provide an evaluation of the food consumed and the user typically has no feedback about their eating behaviours. To overcome these limitations, many studies in the literature have evaluated different healthy quality metrics with again important limitations: some are very simple tools considering dietary recommendations and others are quite complex requiring in-depth analysis of both micro- and macro-nutrient intakes \cite{dietquality1}. Still, there are no satisfactory approaches that consider comprehensive information about the user's eating habits in a user-friendly way and give recommendations to improve the user's health.

On the other hand, over the last decades, a large amount of food-related data have been generated with the use of the internet and smartphones. As a consequence, millions of food images, recipes, cooking videos, and food diaries have exponentially been spread \cite{foodcomputing, bida4}. In addition, new methods such as Automatic Dietary Monitoring (ADM), which uses on-body sensors, have emerged, but existing methods remain inefficient and complex, and require specific sensors \cite{automatic3, automatic4}. Instead of them, user-friendly scenarios based on the acquisition of food images through our personal smartphones can provide reliable feedback on eating behaviours. Several food image databases have been recently released and used for different tasks comprising food retrieval, detection, or recognition, among others \cite{vireo251, UECFood256}. However, despite the popularity of large-scale food image databases, it is noteworthy that there are not any studies to date that provide a useful large-scale image database containing multiple longitudinal eating behaviours, i.e., images of the food consumed along the days for different realistic lifestyle profiles.

This article advances within our interdisciplinary AI4Food project \cite{ai4fooddb}, which intends to break the limits of current personalised health technologies. As a first step, we presented in \cite{romero} our AI4Food-NutritionDB, the only existing database including food images aimed at AI research and personalised nutrition which is balanced and labelled according to a nutritional taxonomy.

The main contributions of this article are:

\begin{itemize}
    \item Proposal of AI4Food-NutritionFW, a framework for the creation of food image datasets according to configurable eating behaviours. The framework simulates a user-friendly and widespread scenario where images are taken using a smartphone. Fig. \ref{fig:confenv} provides a graphical representation of the framework.
    \item The first longitudinal food image dataset existing in the literature aimed at AI-based personalised nutrition research, generated as an example of the use of our proposed AI4Food-NutritionFW. The dataset includes 4,800 different weekly eating behaviours from 15 different profiles and over 1,200 subjects. Note that we incorporate as a novelty the longitudinal aspect with respect to previous works like \cite{romero}. Specifically, we consider profiles that comply with real lifestyles from healthy eating behaviours (according to established knowledge), variable profiles (e.g., eating out, holidays), to unhealthy ones (e.g., excess of fast food and sweets).
    \item Proposal of an automatic healthy index of the subject's eating behaviours using multidimensional metrics based on established guidelines for healthy guidelines.
    \item Public release of our proposed AI4Food-NutritionFW to the research community\footnote{\url{https://github.com/BiDAlab/AI4Food-NutritionFW}}. Being a tunable framework, researchers can now build their own food image datasets based on their own preferences, and research objectives, including more subjects and new profiles.
\end{itemize}

The remainder of the article is organised as follows: state-of-the-art studies related to food recommendation systems and diet quality indicators are presented in Sec. \ref{2}. Sec \ref{3} describes AI4Food-NutritionFW, the framework implemented in our study. Then, the proposed food image dataset and experiments carried out in terms of healthy index evaluation are explained in Sec. \ref{3synthetic} and Sec. \ref{4}, respectively. Then, limitations and challenges of the proposed work are presented in Sec. \ref{5}, and finally, conclusions and future works are described in Sec. \ref{6}.

% % RELATED WORKS SECTION
\section{Related Works}\label{2}

Several national and international organisations have proposed different guidelines to improve public health. For instance, nutritional pyramids indicate how to maintain healthy and balanced eating behaviours along with a balanced physical and mental condition. However, these pyramids are usually specifics of each region and do not consider other external aspects such as different types of food from all over the world, dietary patterns, or lifestyles, among others.

To overcome these aspects, we presented in \cite{romero} an updated version of the food pyramid based on the recommendations of two national organisations - the United States Department of Agriculture (USDA) and the Spanish Society of Community Nutrition (SENC) - and different guidelines from the WHO \cite{pyramid1, pyramid2, dietwho}. It is important to mention that these recommendations are oriented to a daily frequency consumption of basic food products (e.g., fruits or vegetables) and a weekly frequency consumption (e.g., meat and fish). In addition, we also released in \cite{romero} the AI4Food-NutritionDB database, which considers food images and a nutrition taxonomy based on recommendations by national and international organisations.

We describe next state-of-the-art studies related to food recommendation systems and diet quality.

\subsection{Food Recommendation Systems}\label{2a}

State-of-the-art food recommendation systems are normally based on social population groups \cite{diabetes}. Some of them are focused on diabetic individuals, e.g., in \cite{diabetesdietary} the authors implemented a system that was able to create personalised meal plans based on individual requirements such as age, gender, or Body Mass Index (BMI). In addition, they used fuzzy logic to recommend diverse diets including micro and macro nutrients from all the food groups. Similarly, Ali \textit{et al.} proposed in \cite{fuzzy} a system that determined an individual's diet from this health condition. This way the person had to carry several Internet-of-Things (IoT) devices that monitored different biological signals from the body to acquire this data \cite{bida3}.

Diet is one of the most critical aspects of today's society, especially for young people. In particular, Hazman \textit{et al}. proposed a recommendation system that generated healthy meals for children considering some personal characteristics such as age, gender, or health status, among others \cite{childrendietary}. On the opposite extreme, the elderly have today many difficulties when planning a healthy diet or making healthy food choices. Some studies demonstrated that this sector of the population has several social and health problems due to their food intake and, as a consequence, they presented recommendation systems integrated into smartphones and web platforms. For instance, SousChef provides an intuitive environment for older people that generates customised meals from personal information including anthropometric measures, personal preferences, and physical activity \cite{souschef}. NutElCare is another recommendation system that includes healthy guidance supported by experts in nutrition and gerontology \cite{nutelcare}. 

Nutritional misinformation is also present in today's society and can be detrimental to the general population, especially to those people who perform high-intensity physical activity. In \cite{weightlifting}, Tumnark \textit{et al.} implemented an environment that recommended customised diets for weightlifting athletes. They considered a nutritional ontology, a food ontology, a sport profile provided by the user, and different nutritional guidelines for athletes. Finally, many smartphone applications have recently emerged. Most of them calculate nutritional and caloric information from the user's food consumption, for instance, MyFitnessPal\footnote{\url{https://www.myfitnesspal.com}}, MyRealFood\footnote{\url{https://myrealfood.app/}},  or Lifesum\footnote{\url{https://lifesum.com/}}.

To the best of our knowledge, the proposed AI4Food-NutritionFW is the first framework for the creation of food image datasets according to configurable eating behaviours. In particular, we have defined 15 different profiles, considering healthy eating behaviours, variable profiles, and unhealthy ones. Also, since it is a tunable framework, in which researchers can build their own food image datasets, including as many subjects and new profiles as desired.

\subsection{Diet Quality}\label{2b}

The term diet quality emerged in recent decades as the nutritional community needed to somehow measure what a healthy diet looked like. Some studies defined this term as how well an individual is following the dietary recommendations of a particular diet. However, several approaches to measure diet quality exist and the process to standardise a common diet quality index is a complex task, besides the fact that the quality directly depends on how the diet is structured (e.g., Mediterranean diet, vegan diet, gluten-free diet, etc.). Two different types of measurements can be found in the literature in this regard \cite{dietquality1}: \textit{i)} traditional approaches that are usually based on dietary compliance, and \textit{ii)} innovative approaches that consider some aspects such as nutrient or energy density, calories, or even approaches based on the inflammatory potential of the diet. 

Among the different applications developed in this field,  markers for risk assessment, tools for health education, or indicators for health assessment stand out. Specifically, the latter (health assessment indicators), also known as Diet Quality Indices (DQIs), are algorithms to assess the individual's diet based on their eating behaviours \cite{dietquality2}. These algorithms usually utilise both traditional and innovative approaches to create indicators from different dietary methods such as food records, diet histories, or Food Frequency Questionnaires (FFQ) \cite{dietquality3}. For instance, the Healthy Diet Indicator (HDI) is a measure that uses the diet history of an individual \cite{HDI}. In addition, the Healthy Eating Indicator (HEI) is a measure based on nutrients and foods that follows the US Food Guide Pyramid and Dietary Guidelines for Americans \cite{HEI}. However, both indicators have some complex components to automatically calculate such as the percentage of saturated fatty acids or the cholesterol consumed in milligrams, and therefore, some other indicators were created to use them easily. For example, the Mediterranean Lifestyle Index (MEDLIFE) is a  healthy lifestyle index that comprises 28 components of different eating and physical behaviour habits (e.g., the number of servings per week of eggs, fruits, or fish) \cite{MEDLIFE}. 

All previous studies have proposed different metrics for health assessment from a manual user input, which can be a tedious and time-consuming task. AI-based automatic methods have recently gained popularity in food nutrition due to the recent advances in technology and the availability of large-scale databases related to food data. At image level, these two factors have contributed to the improvement of food detection \cite{application_DL, automatic_fd} and recognition systems \cite{food_ai, metwalli}, which are now integrated into various computer programs and mobile applications \cite{ai_food_nutrition}. These AI models have demonstrated impressive performance on some of the most challenging food image databases such as ISIA Food-500 \cite{ISIA} or Food2K \cite{food2k}, specifically in terms of food recognition. However, to the best of our knowledge, there are no studies in the literature that have considered a complete automatic analysis of the food diets or eating behaviours from food pictures. Preliminary approaches have focused instead on calories and volume estimation from food plates to calculate food balances and levels of food consumption during meals \cite{review_applying}. Others have explored different acquisition methods, including food images, embedded cameras, or even microphones to gather food data \cite{allegra2020review, jaswanthi2022hybrid}.

The present work provides an automatic assessment of the individual's eating behaviour from food images, simulating a user-friendly scenario where images of the meal are taken using a smartphone. We also propose an automatic healthy index of the subject's eating behaviour using multidimensional metrics based on the guidelines proposed by international organisations.

% PROPOSED METHODS SECTION
\section{AI4Food-NutritionFW}\label{3}

This section explains in depth the proposed AI4Food-NutritionFW, a framework for the creation of food image datasets with configurable eating behaviour profiles.  Fig. \ref{fig:confenv} graphically summarises our AI4Food-NutritionFW. First, we define the general and food group parameters of the proposed framework to model the different eating behaviour configurations. After that, modifying previous parameters, we can define different profiles. Finally, with the help of the AI4Food-NutritionDB, we can generate datasets with different eating behaviours, select the number of subjects and profiles as desired, or even create new ones. All these tasks are tunable, allowing researchers to build their own datasets and include different configurations. 

\begin{table}[!]
\centering
\caption{Description of the 6 general parameters of the AI4Food-NutritionFW. They define general aspects of how the diet is structured (number of subjects, meals, and main meals), region of the subject, regularity of the diet, and the secondary profile in case of diet irregularities.}
\label{tab:param1}
\begin{tabular}{lc}
\hline
\multicolumn{1}{c}{\textbf{Parameter}} & \textbf{Description}                           \\ \hline
Nº of Subjects                            & Nº of subjects of the profile                           \\
Nº of Meals                            & Nº of meals in a day                           \\
Nº of Main Meals                     & Nº of main meals in a day                    \\
Region                                 & Territorial region of the subject   \\
Regularity                             & Diet's regularity type                         \\
Secondary Profile                         & Type of the second diet \\ \hline
\end{tabular}
\end{table}

\subsection{Eating Behaviour Configuration and Profile Creation}\label{3b}

We initially define the parameters of the AI4Food-NutritionFW to characterise the different eating behaviour profiles. Specifically, we define 2 types of parameters based on general and food-related aspects. First, we consider 6 general parameters that describe the structure of the diet, and second, we provide a parameter for each food group found on the nutritional pyramids (75 in total). Consequently, all 81 parameters considered in this work must be configured to define each eating behaviour profile. These parameters determine the daily and weekly frequency intake of the food products corresponding to the individual's eating behaviour. As guidelines from international organisations are based on daily and weekly recommendations, we determine that all diets must follow a 7-day diet, simulating a natural week. We describe next the details of the general and food group parameters.

\subsubsection{General Parameters}
According to established knowledge, the daily intake should be divided between 3 and 5 meals at regular times. Nevertheless, these meals do not have the same amount of food (normally, breakfast, lunch, and dinner have a higher proportion than between-meals snacks). Therefore, we define two parameters that determine the number of meals and the number of main meals, respectively. In addition, in order to determine the food quantity of each meal, we consider the approach in \cite{romero}, where 7 different types of meals are considered: main meal, appetizer, snack, dessert, side dish, bread, and drinks.

\begin{table}[!]
\centering
\caption{Description of the 75 food group parameters considered in AI4Food-NutritionFW. They define the range of values of the different food groups and products that comprise the individual's eating behaviour.}
\label{tab:param2}
\scalebox{1}{%
\begin{tabular}{lll}
\cline{1-1} \cline{3-3} 
\multicolumn{1}{c}{\textbf{Parameter (I)}}  &  & \multicolumn{1}{c}{\textbf{Parameter (II)}} \\ \cline{1-1} \cline{3-3} 
                                        &  &                                        \\ \cline{1-1} \cline{3-3} 
\textcolor{level1}{\textbf{ Level 1}}                        &  & \textcolor{level4}{\textbf{ Level 4} }                      \\ \cline{1-1} \cline{3-3} 
     $\bullet$ \textbf{Sweet Products}                 &  &$\bullet$ \textbf{Soup \& Stews}       \\ \cline{1-1} \cline{3-3} 
  $\bullet$ \textbf{Fast Food }     &  & $\bullet$ \textbf{Fish and Seafood}         \\\cline{3-3} 
   \quad $ \triangleright$ Vegetable Snacks        &  &   \quad $ \triangleright$ Varied Fish             \\
 \quad $ \triangleright$ Bean Snacks   &  &   \quad $ \triangleright$ Mixed Fish      \\ 
  \quad $ \triangleright$ Other Salty Snacks       &  &    \quad $ \triangleright$ Mollusk           \\ \cline{1-1}
    \quad $ \triangleright$ Pâté                 &  &    \quad $ \triangleright$ Crustacean         \\ \cline{1-1}
\quad $ \triangleright$ Sauce                 &  &   \quad $ \triangleright$ Varied Seafood         \\ \cline{1-1}
\quad $ \triangleright$ Sugary Drinks         &  & \quad $ \triangleright$ Mixed Seafood     \\ \cline{3-3}
\quad $ \triangleright$ Other Drinks          &  &  $\bullet$ \textbf{Beans}            \\ \cline{1-1}  
                                        &  &     \quad $ \triangleright$ Fresh Beans                  \\ \cline{1-1}
\textcolor{level2}{\textbf{ Level 2}}                        &  &  \quad $ \triangleright$ Cooked Beans        \\ \cline{1-1}
$\bullet$ \textbf{Fatty Meat}                    &  & \quad $ \triangleright$ Mixed Beans          \\ \cline{3-3}
\quad $ \triangleright$ Red Meat              &  &  $\bullet$ \textbf{Eggs}          \\  
\quad $ \triangleright$ Breaded Meat          &  &     \quad $ \triangleright$ Eggs (Category)                   \\
\quad $ \triangleright$ Varied Meat           &  &    \quad $ \triangleright$ Mixed Eggs   \\ \cline{3-3} 
\quad $ \triangleright$ Sausage               &  &     $\bullet$\textbf{ Dairy Products}       \\ \cline{1-1} 
\quad $ \triangleright$ Mixed Meat            &  &     \quad $ \triangleright$ Cheese              \\
\quad $ \triangleright$ Dumpling              &  &   \quad $ \triangleright$ Yogurt          \\ \cline{3-3} 
\quad $ \triangleright$ Pie                   &  &     \quad $ \triangleright$  White Meat           \\ 
\quad $ \triangleright$ Stuffed Dough         &  &    \quad $ \triangleright$  Fried Vegetables           \\ \cline{3-3} 
\quad $ \triangleright$ Fried Food            &  & \quad $ \triangleright$ Nut Snacks           \\ \cline{1-1} \cline{3-3} 
\quad $ \triangleright$ Coffee                &  & \quad $ \triangleright$ Rice and Fish        \\
\quad $ \triangleright$ Alcoholic Drinks      &  & \quad $ \triangleright$ Rice and Beans       \\ \cline{1-1}
                                        &  & \quad $ \triangleright$ Sushi                \\ \cline{3-3} 
                        &  &                            \\ \cline{1-1} \cline{3-3} 
\textcolor{level3}{\textbf{Level 3}}                       &  & \textcolor{level5}{\textbf{ Level 5} }                      \\ \cline{1-1} \cline{3-3} 
$\bullet$ \textbf{Sandwich \& Similar} & & $\bullet$ \textbf{Vegetables}                   \\ \cline{1-1}
\quad $ \triangleright$ Fried Seafood         &  & \quad $ \triangleright$ Fresh Vegetables     \\ \cline{1-1}
\quad $ \triangleright$ Fried Beans           &  & \quad $ \triangleright$ Mushrooms            \\ \cline{1-1}
\quad $ \triangleright$ Fried Dairy Products  &  & \quad $ \triangleright$ Cooked Vegetables    \\ \cline{1-1}
\quad  $ \triangleright$ Fried or Breaded Fish  &  & \quad $ \triangleright$ Mixed Vegetables     \\ \cline{1-1}
\quad $ \triangleright$ Toast                 &  & \quad $ \triangleright$ Side Dish Salad      \\ \cline{3-3} 
\quad $ \triangleright$ Other Types of Bread  &  & $\bullet$ \textbf{Fruits}                       \\ \cline{1-1} \cline{3-3} 
 \quad $ \triangleright$ Other Types of Salad &   &                             \\ \cline{1-1} \cline{3-3} 
\quad $ \triangleright$ Rice and Meat         &  & \textcolor{level6}{\textbf{Level 6 (Cereals)}}                       \\ \cline{3-3} 
\quad $ \triangleright$ Meat and Vegetables   &  & $\bullet$ \textbf{Noodle \& Pasta }                \\ \cline{3-3} 
\quad $ \triangleright$ Mixed Food            &  &   $\bullet$ \textbf{Rice}       \\ \cline{1-1}
\quad $ \triangleright$ Vegetable Drinks      &  &   \quad $ \triangleright$ Rice (Category)      \\ \cline{1-1} 
                                        &  &   \quad $ \triangleright$ Mixed Rice           \\ \cline{3-3} 
                                        &  & \quad $ \triangleright$ Bread                \\  \cline{3-3} 
\end{tabular}
}%
\end{table}

\begin{figure*}[!]
      \includegraphics[width=\linewidth]{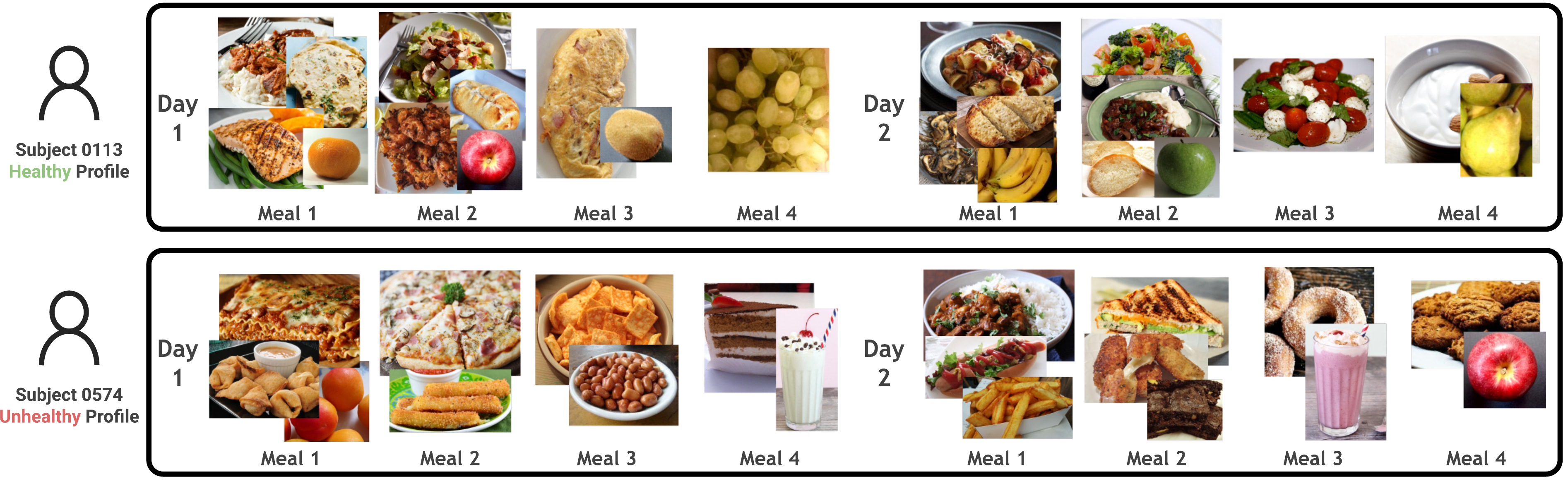}
      \caption{Visual examples from 2 different subjects' eating behaviour. First, subject 0113 (top) has a healthy profile where food products such as rice, fish, or vegetables enrich the diet. The second subject (subject 0574 - bottom), on the contrary, has an unhealthy profile with an abundance of food products from the first nutritional levels such as sweets or fast food. }
    \label{fig:diets}
\end{figure*}

All the general parameters are shown in Table \ref{tab:param1}. The rest of the general parameters are the number of subjects of the corresponding profile, the type of diet (healthy, unhealthy, medium, or variable), the diet regularity that the profile tends to have, the secondary profile (in case of irregularity diet), and the subject's region. This last parameter indicates the world region where the subject is located and consequently, only typical plates of the corresponding region (and international ones) are used from AI4Food-NutritionDB (see Sec. \ref{ai4fooddb}) to feed the dataset. One general region (international) and 6 specific regions are considered: northern America; Latin America and the Caribbean region; Europe; Africa and the West of Asia; Central Asia; and East and Southeast of Asia.

\begin{table}[!]
\centering
\caption{Description of the different eating behaviour profiles considered in the proposed dataset. In \textbf{bold}, we denote the main profiles on which the other profiles of the same type are based on.}
\label{tab:profiles}
\begin{tabular}{ccc}
\hline
\textbf{Type}                                                                 & \textbf{Profile}                  & \textbf{Description}                                                                                                 \\ \hline
                                                                    &  \textbf{1} & \textbf{Follows healthy recommendations}  \\
                                                                              & 1.1      & High alcohol intake \\
\multirow{-2}{*}{\begin{tabular}[c]{@{}c@{}}Healthy\\ Profiles\end{tabular}}   & 1.2                               & \begin{tabular}[c]{@{}c@{}}High intake of food products \\ from nutritional levels 1 and 2\end{tabular}       \\
                                                                              & 1.3      & \begin{tabular}[c]{@{}c@{}}High intake of food products \\ from nutritional level 6\end{tabular}            \\ \hline
                                                                &  \textbf{2} & \begin{tabular}[c]{@{}c@{}}\textbf{High intake of low nutritional} \\ \textbf{quality value foods} \end{tabular}   \\
                                                                              & 2.1     & High alcohol intake           \\
\multirow{-2}{*}{\begin{tabular}[c]{@{}c@{}}Unhealthy\\ Profiles\end{tabular}} & 2.2                               & \begin{tabular}[c]{@{}c@{}}High intake of food products \\ from nutritional level 1\end{tabular}                                         \\
                                                                              & 2.3                               & \begin{tabular}[c]{@{}c@{}}Low intake of \\ healthy food products\end{tabular}                           \\ \hline
                                                                   & \textbf{3}                        & \begin{tabular}[c]{@{}c@{}}\textbf{Balanced diet between healthy}\\ \textbf{and unhealthy food products}\end{tabular}         \\
                                                                              & 3.1                               & \begin{tabular}[c]{@{}c@{}}High intake \\ of meat products\end{tabular}                                   \\
\multirow{-2}{*}{\begin{tabular}[c]{@{}c@{}}Medium\\ Profiles\end{tabular}}    & 3.2                               & \begin{tabular}[c]{@{}c@{}}Highly variable diet\\ during consecutive weeks\end{tabular}                    \\
                                                                              & 3.3                               & \begin{tabular}[c]{@{}c@{}}Similar to 3.2 profile but \\based on a healthy diet\end{tabular}                          \\ \hline
                                                                    & 4.1                        & \begin{tabular}[c]{@{}c@{}}75\% time healthy diet\\ 25\% time unhealthy diet\end{tabular}                           \\
\begin{tabular}[c]{@{}c@{}}Variable\\ Profiles\end{tabular}                    & 4.2                              & \begin{tabular}[c]{@{}c@{}}50\% time healthy diet\\ 50\% time unhealthy diet\end{tabular}                                      \\
                                                                              & 4.3                               & \begin{tabular}[c]{@{}c@{}}75\% time unhealthy diet\\ 25\% time healthy diet\end{tabular}                                      \\ \hline
\end{tabular}
\end{table}

\subsubsection{Food Group Parameters}
These parameters define the intake frequency of the food products during the diet. They are organised into 3 groups, from general to specific ones: \textit{i)} nutritional level (6 parameters), \textit{ii)} category (13 parameters), and \textit{iii)} subcategory (56 parameters). Each group is directly related to the nutritional levels, categories, and subcategories considered in \cite{romero}, respectively. In Table \ref{tab:param2} we describe all the 75 food group parameters considered. Each nutritional level parameter has a unique colour, whereas food category parameters are highlighted in \textbf{bold}. In AI4Food-NutritionFW, at least the nutritional level parameters must be adjusted in order to define a profile. Additionally, we also give the possibility to the research community to generate more specific datasets as desired, adjusting parameters related to the category and subcategory. The parameters also follow a hierarchy in which the most specific group has the highest priority, i.e., if a lower hierarchical level parameter has been set, the framework will consider these parameters and not those of higher levels for the corresponding food products. Similarly, each parameter can be dynamically adjusted by a daily or weekly frequency according to the user's preferences.

Finally, each parameter is then adjusted within a range of values that define a new profile. A final value associated with each parameter is randomly generated between each range in order to differentiate subjects from the same profile, and also to generate different eating behaviours. This task requires a preliminary analysis from a general (comparison among profiles) to a specific point of view to build a reliable dataset. For instance, healthy eating profiles have intake frequencies related to the healthy recommendations and therefore, the parameters will be set according to these guidances. In Sec. \ref{3synthetic} we provide examples of how we adjust these parameters for the 15 eating behaviour profiles considered in the dataset.

After determining the range of values that characterise each profile, the next step computes the parameter values for each unique subject. Random values within the possible ones are then computed to obtain the frequency of each specific parameter, for instance, for a healthy profile with 50 subjects, between 3 and 5 meals, and between 4 and 6 fruits per day, the AI4Food-NutritionFW will be set to each subject a unique value in these ranges. In addition, a balancing process is then executed to have a realistic distribution of values: as parameters are related to food groups and food intake nutritional levels, each frequency is properly distributed among weeks and days.

\subsection{AI4Food-NutritionDB and Food Image Dataset Generation}\label{ai4fooddb}
One of the main contributions of the proposed AI4Food-NutritionFW is to simulate a real environment where people take a picture of food images, providing information related to their eating behaviours. This is done by automatically selecting food images based on the chosen configuration from a large pool of realistic and diverse images: the AI4Food-NutritionDB database \cite{romero}, the first database that considers food images and a nutrition taxonomy based on recommendations by national and international organisations, including four different categorisations: 6 different nutritional levels defined in accordance to the food intake frequency, 19 main food categories (e.g., “Meat”), 73 food subcategories (e.g., “White Meat”), and 893 final food products (e.g., “Chicken”). As the food group parameters of our proposed AI4Food-NutritionFW are directly related to the food image database, a food image of the corresponding group is randomly selected for each meal and day, recreating a realistic situation. The world region is the only restriction considered. Finally, this task is repeated among all the subjects and profiles designed previously and, as a result, a new dataset is generated.

\begin{figure*}[!]
    \begin{center}
      \includegraphics[width=1\linewidth]{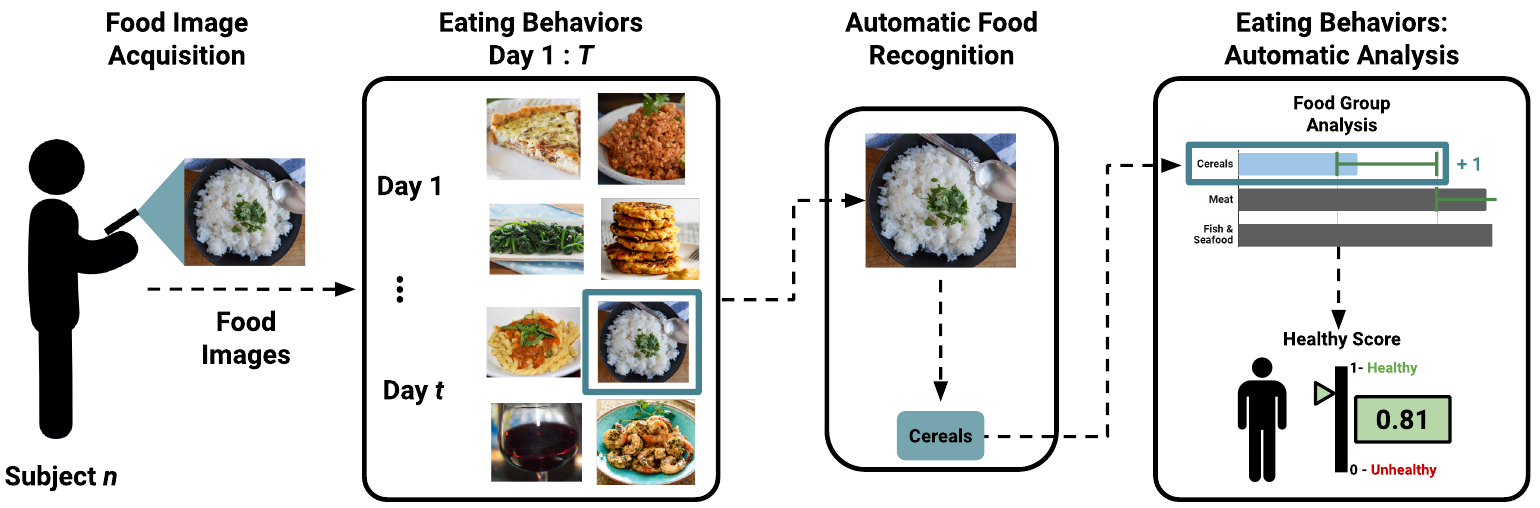}
    \end{center}
      \caption{Food image acquisition and eating behaviour analysis scenario: each food picture taken by the subject is stored over the course of the days and then is passed through an automatic food recognition system. Then, an automatic module analyses all the food consumed in the last week and finally, a Healthy Score that quantifies the subject's eating behaviour is computed.}
    \label{fig:2ndframework}
\end{figure*}

\subsection{Challenges}
Several challenges were faced during the development of our proposed approach. One of the main challenges encountered during the implementation of AI4Food-NutritionFW was to ensure the availability of high-quality and diverse food images that accurately represent different eating behaviours. In this sense, the representation of some meals lacked a real-life aspect. Also, the definition of eating behaviours in a standardized manner proved to be challenging and time-consuming due to the subjective nature of dietary habits. This definition could not have been done properly without the expertise of the nutritionists. 

Furthermore, eating profiles may not fully represent the entire population's eating behaviours. The current implementation of the framework relies on a set of general and food group parameters. To enhance the current framework in the future, several factors, such as cultural, social, or lifestyle, could be integrated into these parameters.

\section{Proposed Food Image Dataset}\label{3synthetic}

To exemplify the use of our proposed AI4Food-NutritionFW, we now create and describe a food image dataset that includes eating behaviours from 15 different profiles and 1,200 subjects. In addition, we define 4 weekly eating behaviours per subject and, as a result, a total of 4,800 different eating behaviours are generated. We consider the nutrition experts' recommendations (from the AI4Food project) and the study developed in \cite{romero} in order to create the different profiles, and to balance the food quantity in each meal, among other aspects. Specifically, we determine 15 profiles that comply with real lifestyles from healthy eating behaviours (i.e., according to established knowledge), variable profiles (e.g., eating out, holidays), to unhealthy ones (e.g., sedentary lifestyle and excess of fast food or sweets) as depicted in Table \ref{tab:profiles}. 

From the 15 profiles created, we define 4 profile types (in \textbf{bold} in Table \ref{tab:profiles}) that denote main profiles on which the other profiles of the same type are based. The Healthy Profile type, in the first place, is directly related to healthy recommendations (high intake frequency of food products from nutritional levels 4, 5, and 6). The Unhealthy Profile type is characterised by high consumption of low nutritional quality food products (products from nutritional levels 1 and 2). For instance, profiles 1.1 and 2.1 are based on profiles 1 and 2, which correspond to healthy and unhealthy profiles, respectively. However, these two profiles are characterised by having a high alcohol intake. The Medium Profile type comprises a balanced diet between healthy and unhealthy profiles. Finally, the Variable Profile type corresponds to those with irregular eating behaviours among different weeks, representing different lifestyles such as eating out or vacations. Profile 4.1, for instance, is a profile that recreates subjects who have  healthy eating behaviours, but for some reason (holidays or similar), they have an entire week of unhealthy eating behaviour (corresponding to the 25\% time of the total diet). 

In order to adjust the different parameters, we first set the general ones. The number of meals is randomly selected between 3 and 5 per day, and between 1 and 3 the number of main meals, depending on each profile. For all of them, we decide to include around 10 subjects per region (including 6 specific regions and 2 international ones) to finally have a total of 80 subjects per profile. For variable profiles, the regularity parameter is set and the secondary profile depends on each scenario. For instance, profile 4.3 is based on a 75\% unhealthy diet and a 25\% healthy one (in this case, the secondary profile parameter is set to healthy). Regarding the food group parameters, we treat each profile individually, e.g., healthy profiles have a higher intake frequency of vegetables per day (between 2 and 4 in most cases) than unhealthy ones (between 1-3 in most cases); on the contrary, unhealthy profiles tend to eat more food products from nutritional level 2 (between 2-7 per week) than healthy ones (0-3 per week).

Some examples of the proposed dataset are shown in Fig. \ref{fig:diets}: subject 0113 (top) has a healthy profile where food products such as rice, fish, or vegetables enrich the diet; the second one (subject 0574 - bottom), on the contrary, has an unhealthy profile with an abundance of food products from the first nutritional levels such as sweets or fast food.

% % EXPERIMENTS AND RESULTS SECTION
\section{Eating behaviours: Automatic Analysis}\label{4}
This section analyses the experiments performed in the proposed food image dataset. Considering different eating behaviours, we first automatically analyse how balanced is each individual's eating behaviour before evaluating its healthy index using multidimensional metrics. It is important to remark that both analyses consider one week's diet since the international organisations' guidelines follow a similar procedure and therefore, we analyse all 4,800 different diets generated (remember that each subject has a 4-week diet).

\begin{figure}[!]
    \centering
    \includegraphics[width=1\linewidth]{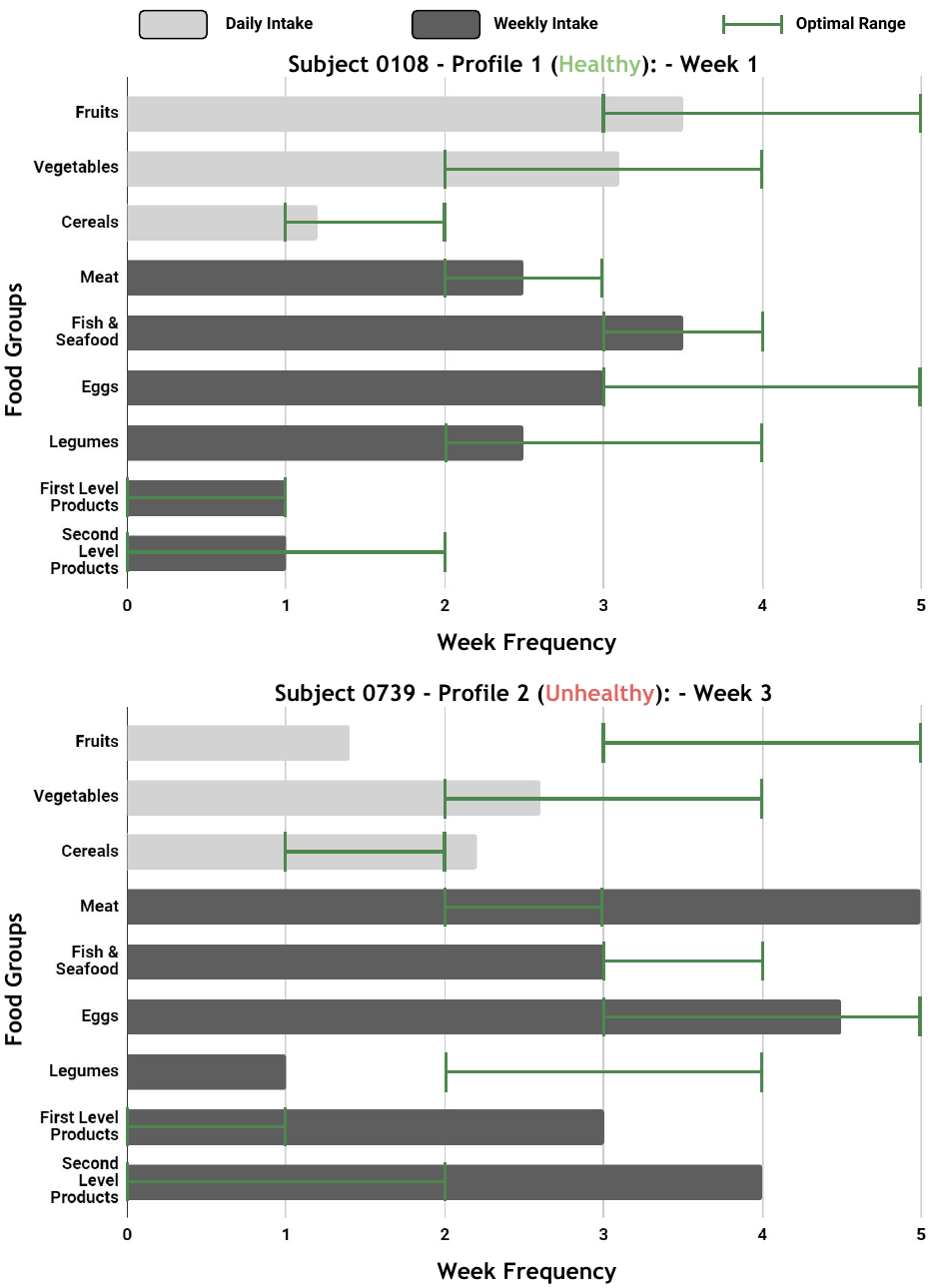}
    \caption{Analysis of two different diets. The first one (subject 0108 - above) corresponds to a subject who leads a healthy eating behaviour, whereas the second subject (subject 0739 - below), corresponds to an unhealthy eating behaviour. Daily average intake frequency groups (in \textcolor{lightgray}{\textbf{light grey}}) and  weekly intake frequency groups (in \textcolor{darkgray}{\textbf{dark grey}})  are shown. In addition, each food group is limited within an optimal range denoted in \textcolor{greenanalysis}{\textbf{green}}.}
    \label{fig:analysis}
\end{figure}

\subsection{Eating Behaviour Analysis: Proposed Method}\label{sec:dietan}
In order to distinguish subjects between healthy and unhealthy eating behaviours, we analyse the subject intake frequency of each main food group over time. As described in Fig. \ref{fig:2ndframework}, each food picture taken by the subject is stored over the course of the days and then passed through an automatic food recognition system. This system automatically detects the main food group. An example of this can be observed in our previous work \cite{romero}, where we trained a state-of-the-art Xception Deep Learning model \cite{xception} using the AI4Food-NutritionDB database. Concretely, the proposed model is based on Convolutional Neural Networks (CNNs) initially pre-trained using the database ImageNet \cite{imagenet}, which consists of over  1M of images from 1,000 distinct classes.

In our model, the last fully-connected layer (FCL) of the original Xception architecture is changed by an FCL with 19 neurons, representing the main food categories. The training process is carried out in two distinct phases. During the first phase, all layers of the network except the FCL ones are frozen in order to train the new FCLs for our specific task. After that, convolutional and FCLs are trained together in order to extract more discriminative features for the task. Both phases are carried out for approximately 50 epochs using Adam optimiser based on binary cross-entropy using a learning rate of 10-3, $\beta1$ and $\beta2$ of 0.9 and 0.999, respectively.  In addition, training and testing are performed with an image size of 224×224. The model with the highest validation accuracy is selected as the best-performing model.  A complete analysis of the food recognition model was performed in \cite{romero}, achieving around 98\% Top-5 accuracy when recognising the main food group.

\begin{figure*}[h]
    \centering
    \includegraphics[width=\linewidth]{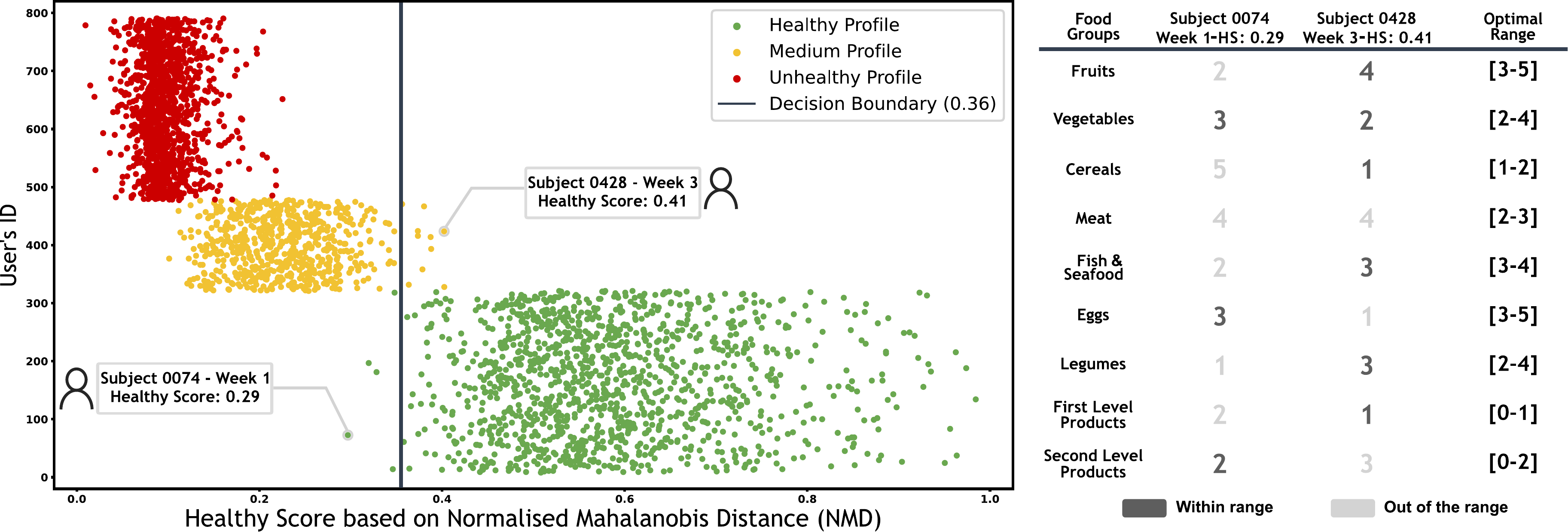}
    \caption{Healthy Score based on the \textit{NMD} applied to the subject's diets from the proposed dataset. Healthy profile subject diets are marked in \textcolor{greenmaha}{\textbf{green}}, unhealthy ones in \textcolor{redmaha}{\textbf{red}}, and medium ones in \textcolor{yellowmaha}{\textbf{orange}}. We also conduct an analysis of two diets that exhibit different eating behaviours compared to their corresponding profiles. Specifically, we compare the food consumption frequencies of these diets with the optimal range. These two diets, belonging to Subject 0074 (week 1) and Subject 0428 (week 3), are visually highlighted on the graph proposed by international organisations. HS = Healthy Score.}
    \label{fig:mahalanobis}
\end{figure*}

Then, after the food recognition module, we can observe our proposed eating behaviour analysis module. This is an automatic module that evaluates all the food consumed by the subject in the last week, and compares it with general recommendations provided by international organisations such as the WHO. In particular, the proposed method considers: \textit{i)} different main food groups, and \textit{ii)} different optimal intake frequency ranges for each of them, according to general recommendations. Particularly, we consider three daily food groups (fruits, vegetables, and cereals), and six weekly food groups (meat, fish and seafood, eggs, legumes, first-level products, and second-level products). Fig. \ref{fig:analysis} provides examples of healthy and unhealthy profiles, including the intake frequency of each food group (daily groups are in \textcolor{lightgray}{\textbf{light grey}} colour whereas weekly groups appear in \textcolor{darkgray}{\textbf{dark grey}}). Also, we highlight in \textcolor{greenanalysis}{green} colour the optimal ranges of each food group according to general recommendations. As we can see in Fig. \ref{fig:analysis}, for the healthy profile, the intake frequency of each food group is close to the optimal range, unlike the unhealthy profile.

\subsection{Healthy Score}
\begin{table}[]
\centering
\caption{Confusion matrix obtained after applying our automatic Healthy Score to the diets considered in the experiment. Healthy diets (1280 total diets) are treated as healthy, and medium (640) and unhealthy (1280) diets are treated as unhealthy.}
\label{tab:confusion}
\begin{tabular}{*{5}{p{13mm}}}
\multicolumn{1}{l}{} & \multicolumn{1}{l}{} & \multicolumn{2}{c}{Predicted Values} &  \\
 &  &  &  &  \\ \cline{3-4}
 & \multicolumn{1}{c|}{} & \multicolumn{1}{c|}{\multirow{3}{*}{\textbf{Healthy}}} & \multicolumn{1}{c|}{\multirow{3}{*}{\textbf{Unhealthy}}} & \multirow{3}{*}{\textbf{Total}} \\
\multicolumn{1}{l}{} & \multicolumn{1}{l|}{} & \multicolumn{1}{c|}{} & \multicolumn{1}{c|}{} &  \\
\multicolumn{1}{l}{} & \multicolumn{1}{l|}{} & \multicolumn{1}{c|}{} & \multicolumn{1}{c|}{} &  \\ \cline{2-4}
\multicolumn{1}{c|}{\multirow{6}{*}{\begin{tabular}[c]{@{}c@{}}Actual\\ Values\end{tabular}}} & \multicolumn{1}{c|}{\multirow{3}{*}{\textbf{Healthy}}} & \multicolumn{1}{c|}{\multirow{3}{*}{1275}} & \multicolumn{1}{c|}{\multirow{3}{*}{5}} & \multirow{3}{*}{1280} \\
\multicolumn{1}{c|}{} & \multicolumn{1}{c|}{} & \multicolumn{1}{c|}{} & \multicolumn{1}{c|}{} &  \\
\multicolumn{1}{c|}{} & \multicolumn{1}{c|}{} & \multicolumn{1}{c|}{} & \multicolumn{1}{c|}{} &  \\ \cline{2-4}
\multicolumn{1}{c|}{} & \multicolumn{1}{c|}{\multirow{3}{*}{\textbf{Unhealthy}}} & \multicolumn{1}{c|}{\multirow{3}{*}{10}} & \multicolumn{1}{c|}{\multirow{3}{*}{1910}} & \multirow{3}{*}{1920} \\
\multicolumn{1}{c|}{} & \multicolumn{1}{c|}{} & \multicolumn{1}{c|}{} & \multicolumn{1}{c|}{} &  \\
\multicolumn{1}{c|}{} & \multicolumn{1}{c|}{} & \multicolumn{1}{c|}{} & \multicolumn{1}{c|}{} &  \\ \cline{2-4}
\end{tabular}
\end{table}

Finally, in order to obtain a score that quantifies the subject's eating behaviour, we compare the similarity of the subject's intake frequency of each food group with the optimal ranges. This is carried out using a Healthy Score based on the Normalised Mahalanobis Distance (\textit{NMD}), where \textit{NMD} is the [0,1] normalised value of the Mahalanobis Distance (\textit{MD}). This distance calculates the similarity between a vector $x_i$ and a set of vectors represented by its mean $\bar{x}$ and Covariance matrix $C_x$ as follows:

\begin{equation}
    \textit{MD}_i = \sqrt{(x_i-\bar{x}) C_x^{-1} (x_i-\bar{x})^T},
\end{equation}

\noindent where $x_i$ is a row vector of size 9 quantifying the 9 specific intakes exemplified in Fig.~\ref{fig:analysis} for subject $i$, $\bar{x}$ is also a row vector of size 9 with the mean for the optimal range $\sigma_j$ for each type of intake $j=1,\ldots,9$, and $C_x=\sigma^T I \sigma$, where $\sigma=[\sigma_1,\ldots,\sigma_9]$ and $I$ is a $9\times9$ identity matrix. In our proposed approach, as implemented in the formulation above, $\bar{x}$ and $C_x$ values are obtained using the optimal ranges of each food group (following healthy recommendations) as this represents the ideal profile of healthy eating behaviour. 

The final Healthy Score is then calculated through the following equation:

\begin{equation}
    \text{\textit{Healthy Score}} = 1 - \text{\textit{NMD}}
\end{equation}

As a result, Healthy Scores close to 1 mean healthy eating behaviours whereas Healthy Scores close to 0 mean unhealthy eating ones. For the automatic analysis, we consider the 3 main profile groups of the proposed dataset (12 different profiles) as depicted in Fig. \ref{fig:mahalanobis}: healthy profiles (in \textcolor{greenmaha}{\textbf{green}}), unhealthy (in \textcolor{redmaha}{\textbf{red}}), and medium ones (in \textcolor{yellowmaha}{\textbf{orange}}). The remaining profile (variable) is not used as it can have a dynamic behaviour (from healthy to unhealthy diet) due to its definition. As a result, 3,200 diets from 960 subjects are analysed. Healthy profile subjects tend to have Healthy Scores between 0.4 and 1, while the trend in unhealthy ones goes to 0. Medium profiles are finally located between them but closer to the unhealthy profile subjects, since they follow a diet combining healthy and unhealthy food products.

Finally, it is possible to classify each eating behaviour by applying thresholds to the values obtained in the Healthy Score. This classification allows for determining whether an eating behaviour is healthy or not. As depicted in Fig. \ref{fig:mahalanobis}, if we set a threshold value of 0.36, Healthy Score values higher than the threshold are treated as healthy diets, whereas values lower than 0.36 are treated as medium and unhealthy ones. 

The performance results achieved in terms of accuracy and sensitivity are 99.53\% and 99.6\%, respectively. These results prove the success of the proposed Healthy Score to automatically assess healthy and unhealthy diets. For completeness, we include in Table \ref{tab:confusion} the confusion matrix, providing an overview of the classification outcomes. As can be seen in Fig. \ref{fig:mahalanobis}, some diets exhibit similarities to non-corresponding profiles, for example, the case of Subject 0074 (week 1) and Subject 0428 (week 3). We analyse in detail these two specific subjects in Fig. \ref{fig:mahalanobis} (right), comparing the actual values of each food category with the optimal range values proposed by international organisations. Values within the range are represented in \textcolor{darkgray}{\textbf{dark grey}}, whereas values outside the range are depicted in \textcolor{lightgray}{\textbf{light grey}}. Notably, Subject 0074 (week 1), corresponding to a healthy profile, tends to have a diet more closely aligned with a medium profile. In this particular case, only vegetables, eggs, and second level products' frequencies are within the range, while others such as cereals or legumes deviate from the optimal values. Conversely, Subject 0428 (week 3) is related to a medium profile, but the corresponding diet is closer to healthy profiles, as indicated by the Healthy Score of 0.41. Similarly, the consumption frequencies of meat, eggs, and second level products in the current diet fall outside the optimal range.

To summarise, this section proves: \textit{i)} the potential of user-friendly scenarios based on food pictures taken by individuals in order to assess eating behaviours, and \textit{ii)} the quality of the proposed dataset for eating behaviours, providing a large range of variability among subjects and profiles.

% % CONCLUSION AND FUTURE WORKS SECTION
\section{Limitations and Future Works}\label{5}

This section describes some of the limitations and challenges of our proposed AI4Food-NutritionFW approach. First, regarding the profile creation module described in Fig. \ref{fig:confenv}, it is crucial for the proper definition of the different eating behaviour profiles and parameters to count on nutritionists. So far, this is a manual and time-consuming process that relies on expert knowledge to ensure accurate adjustments. Nevertheless, future studies could be oriented to automatise the definition of eating behaviours by exploring recent techniques such as Large Language Models (LLMs), e.g., ChatGPT \cite{chatgpt}. 

Focusing on the automatic eating behaviour analysis, we are currently considering a controlled food image acquisition in which the subject must take one picture of each dish. Future approaches could consider more unconstrained scenarios where subjects could capture all the food plates in a single picture (e.g., main meal, appetizer, bread, dessert, etc.). This would require to incorporate a food segmentation module prior to the food recognition stage \cite{myfood, foodpix}.

Another future improvement is related to the Healthy Score proposed in the present study. Currently, our proposed approach is based on data related to eating habits from meals. Future studies may be oriented to include wearable devices in order to acquire other important user habits (e.g., biological and behavioural data). Information related to sleep and physical activities, among others, together with the eating behaviours analysed in the present work, can further enhance the definition of our current Healthy Score. 

Finally, to ensure data privacy preservation, ethical considerations must be thoroughly addressed, especially due to the potential inclusion of sensitive information. This may include metadata (e.g., location, camera settings, software information, etc.) or individuals' appearance in images, as well as user habits across different data points \cite{delgadosurvey}. Protecting the privacy of such data requires careful attention in line with ethical guidelines and regulations \cite{melzi2022overview, delgado2022gaitprivacyon}.

\section{Conclusion}\label{6}

Leading a healthy lifestyle significantly reduces the risk of developing Non-Communicable Diseases (NCD). However, defining and monitoring healthy eating behaviour depends on multiple factors and usually requires the intervention of experts. This article presents and makes public a software implementation of the AI4Food-NutritionFW, a framework for the creation of food image datasets according to configurable eating behaviours. This framework considers several aspects such as the region and lifestyle, and simulates a user-friendly and widespread scenario where food images are taken using a smartphone. In addition, this framework is supported by the AI4Food-NutritionDB, the only existing database that includes food images and a nutritional taxonomy. We also provide a unique food image dataset that includes 4,800 different weekly diets from 15 different profiles (from healthy eating habits to unhealthy ones) and a total of 1,200 subjects. 

We finally assess the healthy index of the subject's eating behaviours using different approaches. First, we analyse the subject's intake frequency of each main food group over time and then we evaluate each eating behaviour using a Healthy Score based on the \textit{NMD}, proving that a healthy eating behaviour can be easily detected.

\section{Funding}
This work has been supported by projects: AI4FOOD-CM (Y2020/TCS6654), FACINGLCOVID-CM (PD2022-004-REACT-EU), INTER-ACTION (PID2021-126521OB-I00 MICINN/FEDER), and HumanCAIC (TED2021-131787B-I00 MICINN).


\end{document}